\documentclass[conference]{IEEEtran}
\IEEEoverridecommandlockouts

\usepackage[inline]{enumitem}
\usepackage{amsmath,amssymb,amsfonts}
\usepackage{algorithmic}
\usepackage{graphicx}
\usepackage{booktabs}
\usepackage{url}
\usepackage{hyperref}
\usepackage{xcolor}
\usepackage{booktabs}
\usepackage[backend=biber,
   style=ieee,
   citestyle=numeric-comp,
   hyperref=true,
   natbib=true,
   maxnames=7,
   minnames=1,
   maxcitenames=1, 
   mincitenames=1,
   giveninits=true,
   hyperref=true,
   sorting=none]{biblatex}
\newcommand{\avgResponseTime}{\bar{T}} 
\newcommand{\cost}{C}
\newcommand{\step}{i} 

\begin{document}

\title{Multi-Objective Deep Reinforcement Learning Optimisation in Autonomous Systems
\\ 
}

\author{
\IEEEauthorblockN{Juan C. Rosero}
\IEEEauthorblockA{School of Computer Science and Statistics\\
Trinity College Dublin\\
Dublin, Ireland \\
roserolj@tcd.ie}
\and
\IEEEauthorblockN{Nicol\'as Cardozo}
\IEEEauthorblockA{DISC \\
Universidad de los Andes\\
Bogot\'a, Colombia \\
n.cardozo@uniandes.edu.co}
\and
\IEEEauthorblockN{Ivana Dusparic}
\IEEEauthorblockA{School of Computer Science and Statistics\\
Trinity College Dublin\\
Dublin, Ireland \\
ivana.dusparic@tcd.ie} 
}

\maketitle

\begin{abstract}

    Reinforcement Learning (RL) is used extensively in Autonomous Systems (AS) as it enables learning at runtime without the need for a model of the environment or predefined actions. However, most applications of RL in AS, such as those based on Q-learning, can only optimize one objective, making it necessary in multi-objective systems to combine multiple objectives in a single objective function with predefined weights. A number of Multi-Objective Reinforcement Learning (MORL) techniques exist but they have mostly been applied in RL benchmarks rather than real-world AS systems. In this work, we use a MORL technique called Deep W-Learning (DWN) and apply it to the Emergent Web Servers exemplar, a self-adaptive server, to find the optimal configuration for runtime performance optimization. We compare DWN to two single-objective optimization implementations:  $\epsilon$-greedy algorithm and Deep Q-Networks. Our initial evaluation shows that DWN optimizes multiple objectives simultaneously with similar results than DQN and $\epsilon$-greedy approaches, having a better performance for some metrics, and avoids issues associated with combining multiple objectives into a single utility function.

\end{abstract}


\section{Introduction}
\label{sec:introduction} 

Autonomous Systems (AS) are designed to operate by continuously adapting to their environment to maintain optimal performance. Self-adaptive Systems (SAS) have the capability to continuously monitor their environment, and self-adapt accordingly, finding more suitable behavior configurations for different environment conditions and situations to maintain an optimal performance~\cite{cheng_software_2009}.

In SAS, predefined actions and action sequences often do not perform optimally due to the dynamic nature of real-world environments, making runtime learning essential for optimizing behavior in dynamic or unpredictable scenarios. 
Reinforcement Learning (RL) is frequently used in these scenarios because it allows systems to adapt at run time. For example, in the domain of web servers and cloud computing, the Fuzzy Q-learning~\cite{jamshidi_fuzzy_2016} RL technique has been used in an automated cloud scaling systems, to minimize the number of response time violations and the amount of resources acquired. Other examples include a combination of both fuzzy Q-learning and SARSA~\cite{arabnejad_comparison_2017}, used to manage auto-scaling, based on the workload, response time and the number of virtual machines, or multi-objective planning and model checking to  maximize revenue, minimize cost, and keep server response time below a threshold on a cloud based load balancing system~\cite{pandey_hybrid_2020}. 

However, most exiting approaches, use a single optimization function, in which multiple objectives are combined at design time into a single one. Truly multi-objective approaches are less frequent. For example, a model based on clustering and situation-driven optimization aims to enhance routing systems by optimizing trip overhead and route cost~\cite{fredericks_planning_2019}. This approach focuses on improving both trip overhead and route cost simultaneously throughout the use of multi-objective Bayesian optimization and multi-objective genetic algorithms. While combining multiple objectives into a single utility function is often used due to its simplicity, Truly Multi-objective Reinforcement Learning (MORL) is crucial for adapting to changing environments and priorities, as adaptability is a lot more challenging to implement on single objective aggregations due to the need to make the combination of multiple objectives at the design phase, making them static.  

In this paper we illustrate the use of RL to implement multi-objective optimization in SAS, using Deep W-learning~\cite{hribar2022deep} (DWN).  DWN is an extension of a multi-objective tabular RL approach called W-learning~\cite{UCAM-CL-TR-362}, which integrates W-learning with neural networks. While tabular W-learning has been successfully applied in multiple examples of SAS (e.g., context-aware adaptation~\cite{cardozo2017peace}, and motorway traffic optimization~\cite{kuvsic2021spatial}), DWN has only been applied to RL benchmarks so far, such as, mountain car and deep sea treasure, but no real world application has been implemented. 

The contribution of this paper is the first implementation of the DWN Multi-Objective Deep Reinforcement Learning algorithm in an autonomous system. Specifically, we apply DWN to the Emergent Web Server (EWS) self-adaptive exemplar~\cite{filho2022emergent}, which dynamically transitions server configurations to find the optimal setup. We evaluate our multi-objective approach against the baseline $\epsilon$-greedy algorithm in EWS and a Deep Q-Networks (DQN) approach, optimizing for average response time and configuration cost. The full implementations of both DWN and DQN applied to EWS are available in our repository.\footnote{\url{https://github.com/JuanK120/RL_EWS}} Our initial results show DWN outperforms both $\epsilon$-greedy and DQN in average response time, though with slightly higher cost variance compared to $\epsilon$-greedy, which performed best for cost.

The rest of the paper is organized as follows. Section~\ref{sec:MO-Optim} covers background and state of the art in multi-objective optimization in autonomous systems. Section~\ref{sec:DWL} presents DWN, laying the foundations for our proposed implementation, while section~\ref{sec:DWNImp} describes our implementation and test environment (EWS). Section~\ref{sec:Comp} presents the results of our performance tests and discusses the findings, while final section concludes the paper and outlines possible future work.


\section{Multi-Objective Optimisation in Autonomous Systems}
\label{sec:MO-Optim}

Multiple objectives in AS can be optimized using various techniques and algorithms. For instance, genetic algorithms have been applied to multi-objective scheduling of autonomous parking robots~\cite{chen_multiobjective_2021} and autonomous path planning for re-configurable tiling robots~\cite{cheng_multi-objective_2020}. CrowdNav~\cite{fredericks_planning_2019} employs both genetic algorithms and Bayesian optimization for bus routing, optimizing trip overhead and cost. Investigations into non-adaptive mixed and adaptive strategies have been conducted to achieve Pareto efficiency in self-organizing camera networks~\cite{lewis_static_2015}. 

RL has seen extensive use for optimization in AS and SAS, such as using fuzzy Q-learning and SARSA~\cite{arabnejad_comparison_2017} to optimize workload, response time, and the number of virtual machines in auto-scaling for cloud systems. Q-learning has also been used for multi-objective path planning in unmanned aerial vehicles~\cite{zhang_multi-objective_2022, ramezani_human-centric_2024}.

The RL examples mentioned above adapt single-objective RL methods to multiple objectives, but true MORL applications in real-world scenarios are rarer. These include hybrid multi-objective reinforcement learning and deep neural networks for radio communications~\cite{ferreira_multi-objective_2017}, constrained MORL for automated vehicle suspension control~\cite{wang_enhancing_2024}, and personalized decision-making in autonomous driving~\cite{he_toward_2023}.

These examples demonstrate the necessity of multi-objective optimization for enhancing performance and efficiency in autonomous systems, especially in balancing conflicting areas such as response time and error rate. Most existing approaches focus on single-objective functions, but true MORL is crucial for adapting to changing objectives and priorities, which is challenging for single-objective aggregation methods.

In the next section, we present the details of our MORL, which flexibly optimizes individual objectives and then combines them for a truly multi-objective approach.
    

\section{Distributed W-Learning for Optimisation in Autonomous Systems}
\label{sec:DWL} 

This section introduces key concepts to our solution. First, Deep Q-Learning (DQN)~\cite{article_AtariDRL}, an RL method that extends regular Q-learning to use deep learning. Second, we introduce Deep W-Networks~\cite{hribar2022deep}, a DQN extension that implements MORL.

\subsection{Deep Q-learning}

The goal of RL is to find the optimal policy $\pi^*$ for an environment That is characterized by a Markov Decision Process (MDP). An MDP comprises a state space $S$, an action space $A$, a reward function $R$, and state transition probability $P$. The optimal policy $\pi^*$ maximizes long-term rewards over short-term rewards. The DQN algorithm, extends Q-learning using deep learning. In Q-learning, the agent interacts with the environment, receiving a reward $r$ for an action $a$, aiming to estimate the action-value function $Q(s, a)$. The RL algorithm iteratively updates Q-values using the Bellman equation:
    \[
    Q_{i+1}(s, a) = \mathbb{E}_{s^{\prime} \sim S} [r = \gamma \max_{a^{\prime}} Q^*(s^{\prime}, a^{\prime}) | s, a],
    \]
where $\gamma$ is the discount factor. These values converge to the optimal value $Q^*$ as $i \to \infty$.
    
However, exhaustive state-space exploration is computationally impractical. Instead, function approximators like artificial neural networks (ANN) are used. In DQN, an ANN with weights $\theta$ approximates the Q-network, updated by minimizing the loss:
    
    \[
    L_i(\theta_i) = \mathbb{E}_{s,a \sim \rho(\cdot); s^{\prime} \sim S} \left[(y_i - Q(s, a; \theta_i))^2 \right],
    \]
    
where $y_i = \mathbb{E}_{s \sim S} [r + \gamma \max_{a^{\prime}} Q(s^{\prime}, a^{\prime}; \theta_{i-1}) | s, a]$. Gradient descent optimizes this loss, but training can be unstable. To address this, DQN uses experience replay and separate target networks. Experience replay stores experiences $(s, a, s^{\prime}, r)$, sampling them uniformly during training. Prioritized Experience Replay (PER) improves this by prioritizing important experiences, with sampling probability $P(i) = \frac{p_i^\zeta}{\sum_k p_k^\zeta}$, where $\zeta \in [0, 1]$. Target networks stabilize training by providing consistent Q-value estimates.      

\subsection{Deep W-Networks} 

In DWN, the goal is to adapt DQN for optimizing multiple objectives simultaneously, specifically average response time ($\avgResponseTime$) and configuration cost ($\cost$). This is done by optimizing objectives separately with different policies that suggest potentially different actions (selecting a configuration in EWS). A mechanism then chooses the best action for the situation.

Each policy has two DQN networks: one for Q-values and one for W-values, with two replay memories for Q-networks and W-networks. DWN is defined by the tuple $\langle N, S, A, R, \Pi \rangle$. Here, $N$ is the number of policies, $S$ is the state space, $A$ is the set of actions, $R = \{R_1, \ldots, R_N\}$ is the set of reward functions, and $\Pi = \{\pi_1, \ldots, \pi_N\}$ is the set of policies. At each decision epoch $t$, policies observe state $s(t) \in S$ and suggest actions. The agent selects the best action, $a_j(t)$, from these suggestions.The objective is to balance multiple objectives using W-learning, with each objective represented by a Q-learning policy suggesting a greedy action $a_i(t)$. Conflicts are resolved by learning W-values for each state and selecting the action from the policy with the highest $W_i(t)$:

    \[
    W_j(t) = \max \{W_1(t), \ldots, W_N(t)\}.
    \]
The W-values are updated similarly to Q-values but excluding the selected policy:
    
    {\begin{align*}
        W_i(t) &\leftarrow (1 - \alpha)W_i(t) + \alpha [ Q(s(t), a_j(t)) - (R_i(t) + \\
        &\gamma \max_{a_i(t+1) \in A} Q(s(t+1), a_i(t+1)))].
    \end{align*}}

\begin{figure}[h]
        \centering
        \includegraphics[width=0.7\columnwidth]{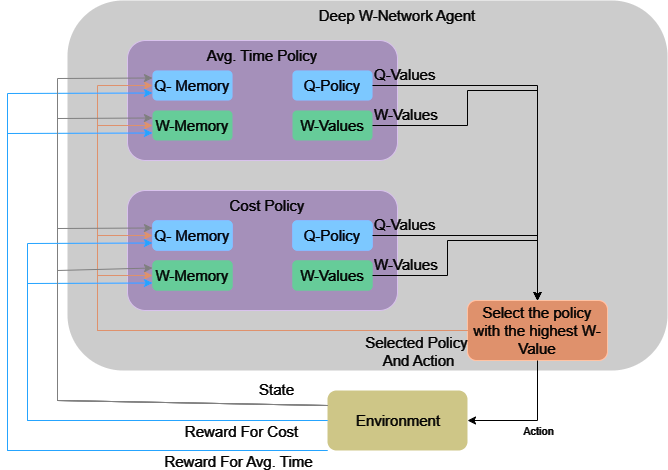}
        \caption{Working of a DWN network}
        \label{fig:DwnNEt}
    \end{figure}

The \(\epsilon\)-greedy strategy facilitates exploration, requiring at least \(K\) experiences in replay memory before training starts. During training, policy networks are updated first, followed by W-networks after a delay to ensure stability. This supports diverse policy learning and effective multi-objective optimization. Figure \ref{fig:DwnNEt} illustrates the DWN architecture, where actions initially train individual DQNs for each objective, and W-learning then determines the optimal action for the state based on learned weights associated with each objective.

\section{Implementation}
\label{sec:DWNImp}

In this section we evaluate our implementation to optimize server configurations in EWS,\footnote{\url{https://github.com/EGAlberts/pyews}} an exemplar to implement online learning in self-adaptive systems, this server has the capability of changing configurations at run time. 

\subsection{Emergent web server}

Emergent Web Server (EWS)~\cite{filho2022emergent} is a self-adaptive server capable of changing configurations at runtime. It is implemented in the Dana language. \footnote{https://www.projectdana.com} 
A server configuration in EWS consists of small, reusable components, each implementing a specific interface with defined functions. These components can be swapped at runtime to adapt the server's behavior to changing conditions. The base components that compose different configurations include:
\begin{itemize*}
    \item Request handlers
    \item HTTP processing modules
    \item Additional required modules, such as compression algorithms and cache strategies.
\end{itemize*}

In baseline EWS, these components enable a total of 42 different configurations, formed from 2 request handlers, 4 HTTP processing modules, 2 compression algorithms, and 4 cache strategies. Furthermore, EWS includes mechanisms for obtaining and utilizing response time as a performance metric.
To facilitate interactions with EWS, a Python module is provided, which not only allows for the implementation of these configurations in Python scripts but also includes an $\epsilon$-greedy algorithm as a baseline optimization technique for response time. We utilize this Python module and the 42 configurations as the foundation for our implementation.

\subsection{Scenario Description}

We compare our DWN implementation to a modified version of the baseline epsilon-greedy ($\epsilon$-greedy) algorithm implemented in EWS, and an implementation of DQN, both using scalar addition to combine multiple objectives. The goal is to optimize server performance across two metrics:
\begin{enumerate*}[label=(\arabic*)]
\item Average response time to requests $\avgResponseTime$ during a three-second data collection window (step $\step$).
\item Cost $\cost$, associated with using a specific configuration.
\end{enumerate*}

The cost metric was not originally part of the algorithms. The cost metric was introduced and randomly generated, with manually created scenarios where cost conflicts with response time to illustrate the benefits of a multi-objective approach.

The goal of our optimization is to minimize both $\avgResponseTime$ and $\cost$. Originally, the $\epsilon$-greedy algorithm considers only one objective. Since both $\avgResponseTime$ and $\cost$ are within a similar value range (from zero to one), we used an aggregated reward value for each $\step$ in the $\epsilon$-greedy algorithm. We applied the same aggregated reward to DQN, as it shares the same limitation. Additionally, to avoid configurations being optimal for one objective but not for the other, we introduced a measure to balance the optimization. Our individual rewards for each objective are calculated for every $\step$:

    \begin{align}
        & \text{if } \avgResponseTime_{\step} > 0.5 \text{ then } r_{\avgResponseTime_{\step}} = 2 \text{ else } r_{\avgResponseTime_{\step}} = \avgResponseTime_{\step} \label{eq:rwrdAvgTime} \\
        & \text{if } \cost_{\step} > 0.5 \text{ then } r_{\cost_{\step}} = 2
        \text{ else } r_{\cost_{\step}} = \cost_{\step} \label{eq:rwrdCost}
    \end{align} 

A notable concern regarding our comparison is the potential bias introduced by the rewards function defined in Equations \ref{eq:rwrdAvgTime} and \ref{eq:rwrdCost}. The choice to apply a penalty when either the average response time or cost exceeds 0.5 may indeed overemphasize these objectives. This weighting strategy was implemented to prevent the agent from excessively favoring one objective over the other during training.

The global rewards for DQN and $\epsilon$-greedy are calculated as follows:

    \begin{align}
        &  r_{\step} = - r_{\avgResponseTime_{\step}} - r_{\cost_{\step}} \label{eq:rwrdDqnEgreedy} 
    \end{align}

For DWN, the global reward is a tuple composed of both values:

    \begin{align}
        &  r_{\step} = \langle - r_{\avgResponseTime_{\step}} , - r_{\cost_{\step}}\rangle \label{eq:rwrdDWN} 
    \end{align}

Regarding the optimization, once the individual rewards are computed using equations \ref{eq:rwrdAvgTime} and \ref{eq:rwrdCost}, both DQN and $\epsilon$-greedy optimize a combined reward, as expressed in Eq. \ref{eq:rwrdDqnEgreedy}. Optimization in DWN consists on 2 individual DQNs that are trained with the separate components of the tuple defined in Eq. \ref{eq:rwrdDWN}, which represent the individual rewards for $\avgResponseTime$ and $\cost$. These individual DQNs are subsequently integrated using the DWN network.

For the hyper-parameters, we set $\epsilon = 0.0001$ for the baseline $\epsilon$-greedy algorithm. hyper-parameters for DQN can be found in Table \ref{tab:HParamsDQN}, and for DWN in Table \ref{tab:HParamsDWN}.

    \begin{table}[htbp] 
        \centering
        \caption{Hyperparameters for DQN}
        \begin{tabular}{c|c || c|c}
            \toprule
            \textbf{Hyperparameter} & \textbf{Value} & \textbf{Hyperparameter} & \textbf{Value} \\
            \midrule
            Optimizer & Adam & Memory size ($M$) & $10^6$ \\
            Batch size ($K$) & 64 & Learning rate ($\alpha$) & 0.01 \\
            Start epsilon ($\epsilon_{start}$) & 0.99 & Min. epsilon ($\epsilon_{min}$) & 0.001 \\
            Epsilon decay ($\epsilon_{decay}$) & 0.99 & Discount factor ($\gamma$) & 0.99 \\
            \bottomrule
        \end{tabular}
        \label{tab:HParamsDQN}
    \end{table}

    \begin{table}[htbp]
        \centering
        \caption{Hyperparameters for DWN}    
        \begin{tabular}{c|c || c|c}
            \toprule
            \textbf{Hyperparameter} & \textbf{Value} & \textbf{Hyperparameter} & \textbf{Value} \\
            \midrule
            Q Optimizer & Adam & W Optimizer & Adam \\
            Memory size ($M$) & $10^6$ & Batch size ($K$) & 64 \\
            Learning rate ($\alpha$) & 0.01 & Discount factor ($\gamma$) & 0.99 \\
            Q Epsilon ($\epsilon^{Q}$) & 0.99 & Min. Q epsilon ($\epsilon^{Q}_{min}$) & 0.001 \\
            Q Eps. decay ($\epsilon^{Q}_{decay}$) & 0.99 & W Epsilon ($\epsilon^{W}$) & 0.99 \\
            Min. W epsilon ($\epsilon^{W}_{min}$) & 0.001 & W Eps. decay ($\epsilon^{W}_{decay}$) & 0.99 \\
            \bottomrule
        \end{tabular}
        \label{tab:HParamsDWN}
    \end{table}

These values for hyperparameters were chosen because they are among the most common in reinforcement learning literature and provide a good starting point. However, further optimization could potentially improve performance and will be considered in future work.
    
Our test is designed as follows: We start EWS and initialize a client script that sends requests sequentially, storing the response time. Then, we execute our optimization algorithms. During each 3 second collection window $\step$, we gather response times for received requests and the cost $r_{\cost}$ of the chosen configuration. We compute $r_{\avgResponseTime}$ from the collected data. The algorithms test different configurations in EWS, aiming to minimize both the average response times and the cost for each $\step$.

\section{Experimental results and analysis}
\label{sec:Comp}

We conducted the experiment 8 times, storing data in separate data-frames. Each run involved randomly shuffling configurations to ensure algorithms explored and found optimal solutions. This tests were done on a windows machine with an Intel Core 7 processor 155H, 16GB Ram memory, and RTX 4060 GPU. Detailed results are available in the implementation repository's results folder:\footnote{\url{https://github.com/JuanK120/RL_EWS/tree/master/results}}
    
    \begin{figure}[htbp]
        \centering
        \includegraphics[width=0.7\columnwidth]{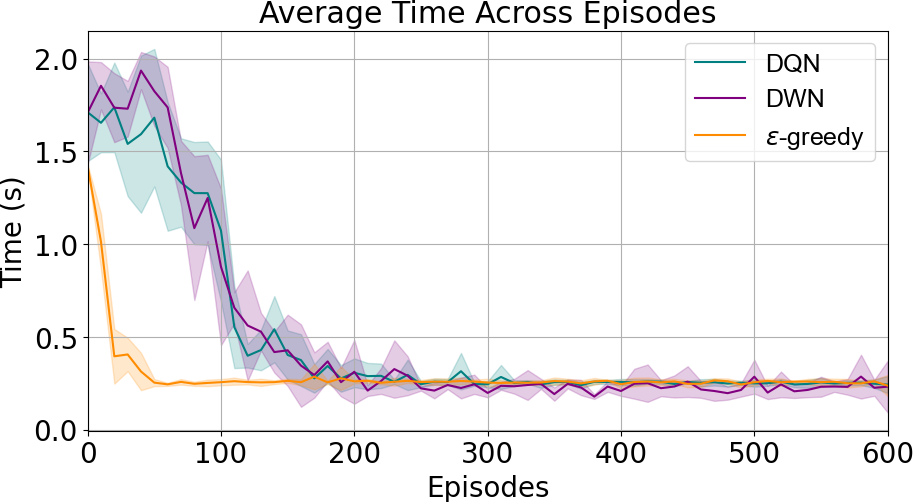}
        \caption{Comparison of the evolution of avg. response time}
        \label{fig:Avg.Plot}
    \end{figure}
    \begin{figure}[htbp]
        \centering
        \includegraphics[width=0.7\columnwidth]{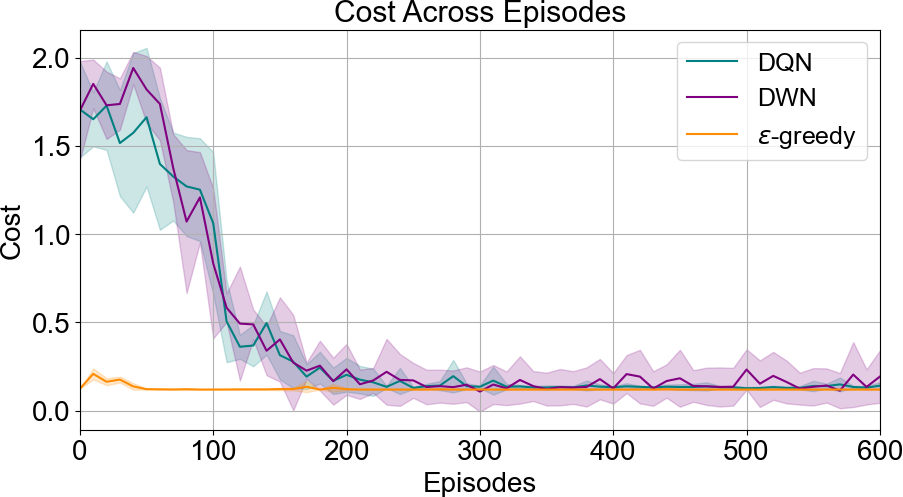}
        \caption{Comparison of the evolution of Cost}
        \label{fig:Cost.Plot}
    \end{figure}

Figures \ref{fig:Avg.Plot} and \ref{fig:Cost.Plot}, respectively illustrate the evolution of average response time and cost. Notably, $\epsilon$-greedy, DWN, and DQN show similar optimization trends, converging around 64 steps and fully by 200 steps. The quicker convergence of $\epsilon$-greedy stems from its early emphasis on exploiting known configurations, while DWN and DQN first explore various options for 64 steps before optimizing based on gathered data.

Additionally, Table \ref{tab:metricsComparisson} presents the average response time  and cost, for the last hundred steps of the episode. The table provides insight about the performance of the DWN, DQN, and $\epsilon$-greedy algorithms during the latter stages of execution, comparing them to one another. We can observe that DWN provides a better $\avgResponseTime$ in comparison to DQN and DWN, while having a  slightly higher $\cost$. However, due to the alternating of configurations from both DWN and DQN, the deviations resulting from that could possibly make the results less robust than they could appear at first sight due to results being within the uncertainty bounds.

    \begin{table}[htbp]
        \centering
        \caption{Comparison of metrics across different algorithms}
        \begin{tabular}{c | c | c}
            \toprule
            \textbf{ } & \textbf{$\avgResponseTime$} & \textbf{$\cost$} \\
            \midrule
            DWN & \textbf{0.2362 $\pm$ 0.1196} & 0.1679 $\pm$ 0.1420 \\
            DQN & 0.2480 $\pm$ 0.0410 & 0.1387 $\pm$ 0.0187 \\
            $\epsilon$-greedy & 0.2523 $\pm$ 0.0554 & \textbf{0.1191 $\pm$ 0.0000} \\
            \midrule
            DWN - Policy Avg. time & \textbf{0.2138 $\pm$ 0.0489} & 0.2267 $\pm$ 0.1049 \\
            DWN - Policy Cost & 0.2738 $\pm$ 0.0507 & \textbf{0.0026 $\pm$ 0.0000} \\
            \bottomrule
        \end{tabular}
        \label{tab:metricsComparisson}
    \end{table}    
  
DWN slightly outperformed the other two algorithms in average response time and demonstrated similar cost performance, albeit with greater instability, as indicated by the shaded region. Unlike $\epsilon$-greedy, which selected a single best configuration, DWN evaluated various candidate configurations with different costs and trade-offs between cost and average time, leading to larger deviations but allowing for greater flexibility.

DQN on the other hand, possibly due to the use of same reward system of $\epsilon$-greedy algorithm, a linear aggregation of the metrics, reached a similar result too, meaning a single configuration identified as the overall best one and sticking to that one most of the time, resulting in a much smaller deviation than DWN, although still alternating from time to time, demonstrating a bigger degree of flexibility than $\epsilon-greedy$. $\epsilon$-greedy chose one option and completely stuck to it, with little to no configuration changes after about the 70Th step.

    \begin{figure}[htbp]
        \centering
        \includegraphics[width=0.7\columnwidth]{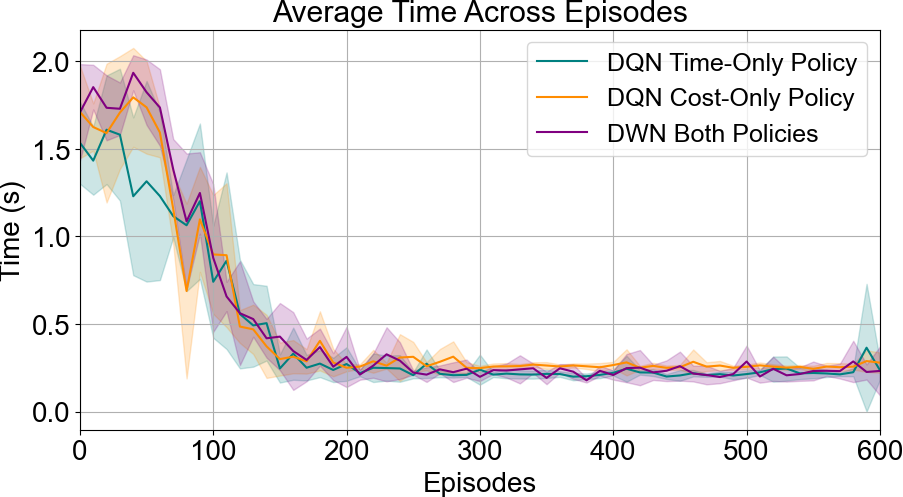}
        \caption{Behavior of DWN in avg. response time}
        \label{fig:avg_dwn_policies.Plot}
    \end{figure}
    \begin{figure}[htbp]
        \centering
        \includegraphics[width=0.7\columnwidth]{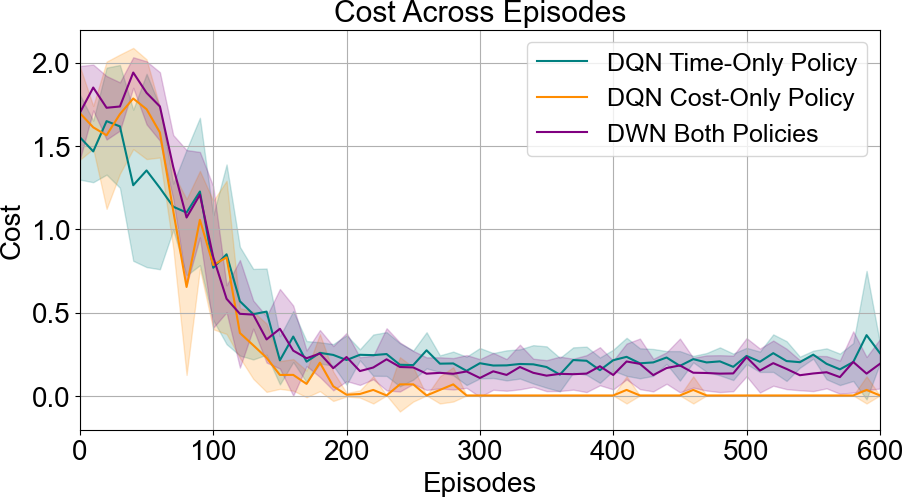}
        \caption{Behavior of DWN in cost}
        \label{fig:Cost_dwn_policies.Plot}
    \end{figure}

Separately, we also conducted an experiment with DWN by separating both policies into distinct DQN networks. Figures \ref{fig:avg_dwn_policies.Plot} and \ref{fig:Cost_dwn_policies.Plot} show the performance for the policies of DWN and the complete DWN algorithm integrating w-learning, for $\avgResponseTime$ and $\cost$ respectively. The results clearly demonstrate that the Cost-only and Time-only policies effectively optimize their respective objectives, outperforming other algorithms. This is also evident in Table \ref{tab:metricsComparisson}, where the last hundred steps show both separate policies achieving the best values for their respective objectives compared to DQN, $\epsilon$-greedy, and DWN. Additionally, DWN finds a balance between both objectives, achieving the best performance for $\avgResponseTime$ without significantly compromising cost, only about $0.05$ higher than $\epsilon$-greedy, which was the best performer for $\cost$.

Overall, DWN showed superior average response time despite higher variability and slightly increased overall cost, owing to its strategy of alternating between multiple near-optimal configurations. In contrast, DQN, using a reward system akin to $\epsilon$-greedy, focused more on identifying a single optimal configuration, leading to lower variability and potentially less adaptability. The $\epsilon$-greedy algorithm's quick selection of a single configuration may provide stability but could be less flexible due to its approach.

\section{Conclusion and Future Work}
\label{sec:ConclusionsFtrWork}

This paper presents the implementation and evaluation of a Multi-Objective Reinforcement Learning algorithm using Deep W-Networks to optimize configurations in the Emergent Web Server. Our primary goal was to demonstrate the first application of RL in an AS, as DWN has previously only been tested in RL toy example benchmarks. We used DWN in EWS to minimize two performance metrics: average response time and cost of the available configurations. We compared DWN's performance with a modified version of the baseline $\epsilon$-greedy algorithm and a Deep Q-Network approach, both of which used an aggregated value to optimize $\avgResponseTime$ and $\cost$.
    
Our experimental results demonstrate that both DQN and DWN perform comparably to the $\epsilon$-greedy algorithm in these metrics. DWN achieved the lowest average response time ($\avgResponseTime$), outperforming DQN by $4.75\%$ and $\epsilon$-greedy by $6.43\%$. However, this improvement came at a higher cost, with DWN's cost being $21.41\%$ higher than DQN and $40.88\%$ higher than $\epsilon$-greedy.Notably, these percentage differences are compelling but must be interpreted with caution due to the error bounds and the deviations resulting from alternating configurations.

DWN achieved a better trade-off between response time and cost compared to DQN. While DQN's average response time was only $1.7\%$ lower than the baseline $\epsilon$-greedy, its cost was $14.13\%$ higher. For DQN to achieve a trade-off similar to DWN, its cost would need to be approximately $49.13\%$ higher, whereas DWN's cost increase was only $40.88\%$. Another observation was the strategies the algorithms followed. The $\epsilon$-greedy algorithm tended to find an optimal configuration and stick with it, while DQN showed a middle ground, mostly sticking with one optimal configuration but occasionally changing. DWN, on the other hand, sought a compromise between the DQN agents optimizing separate objectives within the algorithm.

For future work, we propose more extensive analysis involving multiple metrics, such as resource consumption, which are difficult to combine into a single objective. This is currently unfeasible due to data collection limitations. Testing performance in scenarios where DQN or $\epsilon$-greedy struggle with multiple objectives would highlight another advantage of DWN, which doesn't require aggregating values for multi-objective optimization, as well as studying further tuning of hyperparameters to optimize performance. Additionally, we propose implementing combinations and mutations between EWS configurations to better optimize multiple objectives. Finally, we are exploring integrating more complex multi-objective optimization frameworks, such as ComInA\cite{cardozo2020learning}, that even presents an application to a bus routing systems.

\section*{Acknowledgment}


 This publication has been supported in part by the Science Foundation Ireland under Grant number 18/CRT/6223 For the purpose of Open Access, the author has applied a CC BY public copyright license to any Author Accepted Manuscript version arising from this submission.

%
%
\printbibliography

@inproceedings{hribar2022deep,
  title={Deep W-Networks: Solving Multi-Objective Optimisation Problems With Deep Reinforcement Learning},
  author={Jernej Hribar and Luke Hackett and Ivana Dusparic},
  booktitle={International Conference on Agents and Artificial Intelligence},
  year={2022}, 
}

@inproceedings{filho2022emergent,
  title={Emergent web server: An exemplar to explore online learning in compositional self-adaptive systems},
  author={Filho, Roberto Rodrigues and Alberts, Elvin and Gerostathopoulos, Ilias and Porter, Barry and Costa, F{\'a}bio M},
  author={Filho, Roberto Rodrigues and Alberts, Elvin and Gerostathopoulos, Ilias and Porter, Barry and Costa, F{\'a}bio M},
  booktitle={Proceedings of the 17th Symposium on Software Engineering for Adaptive and Self-Managing Systems},
  pages={36--42},
  year={2022}
}

@inproceedings{cardozo2020learning,
  title={Learning run-time compositions of interacting adaptations},
  author={Cardozo, Nicol{\'a}s and Dusparic, Ivana},
  booktitle={Proceedings of the IEEE/ACM 15th international symposium on software engineering for adaptive and self-managing systems},
  pages={108--114},
  year={2020}
}

@inproceedings{fredericks_planning_2019, 
	title = {Planning as {Optimization}: {Dynamically} {Discovering} {Optimal} {Configurations} for {Runtime} {Situations}},
	copyright = {https://ieeexplore.ieee.org/Xplorehelp/downloads/license-information/IEEE.html}, 
	shorttitle = {Planning as {Optimization}},
	language = {en},
	booktitle = {2019 {IEEE} 13th {International} {Conference} on {Self}-{Adaptive} and {Self}-{Organizing} {Systems} ({SASO})},
	publisher = {IEEE},
	author = {Fredericks, Erik M. and Gerostathopoulos, Ilias and Krupitzer, Christian and Vogel, Thomas},
	month = jun,
	year = {2019},
	pages = {1--10},
}

@incollection{cheng_software_2009,
	address = {Berlin, Heidelberg},
	title = {Software {Engineering} for {Self}-{Adaptive} {Systems}: {A} {Research} {Roadmap}},
	volume = {5525}, 
	shorttitle = {Software {Engineering} for {Self}-{Adaptive} {Systems}},
	language = {en},
	booktitle = {Software {Engineering} for {Self}-{Adaptive} {Systems}},
	publisher = {Springer Berlin Heidelberg},
	author = {Cheng, Betty H. C. and De Lemos, Rogério and Giese, Holger and Inverardi et al.},
	editor = {Cheng, Betty H. C. and De Lemos, Rogério and Giese, Holger and Inverardi, Paola and Magee, Jeff},
	year = {2009}, 
	note = {Series Title: Lecture Notes in Computer Science},
	pages = {1--26},
}

@inproceedings{jamshidi_fuzzy_2016, 
	title = {Fuzzy {Self}-{Learning} {Controllers} for {Elasticity} {Management} in {Dynamic} {Cloud} {Architectures}},
	language = {en},
	booktitle = {2016 12th {International} {ACM} {SIGSOFT} {Conference} on {Quality} of {Software} {Architectures} ({QoSA})},
	publisher = {IEEE},
	author = {Jamshidi, Pooyan and Sharifloo, Amir and Pahl, Claus and Arabnejad, Hamid and Metzger, Andreas and Estrada, Giovani},
	month = apr,
	year = {2016},
	pages = {70--79},
}

@inproceedings{arabnejad_comparison_2017, 
	title = {A {Comparison} of {Reinforcement} {Learning} {Techniques} for {Fuzzy} {Cloud} {Auto}-{Scaling}},
	language = {en},
	booktitle = {2017 17th {IEEE}/{ACM} {International} {Symposium} on {Cluster}, {Cloud} and {Grid} {Computing} ({CCGRID})},
	publisher = {IEEE},
	author = {Arabnejad, Hamid and Pahl, Claus and Jamshidi, Pooyan and Estrada, Giovani},
	month = may,
	year = {2017},
	pages = {64--73},
}

@article{lewis_static_2015,
	title = {Static, {Dynamic}, and {Adaptive} {Heterogeneity} in {Distributed} {Smart} {Camera} {Networks}},
	volume = {10},
	issn = {1556-4665, 1556-4703},
	language = {en},
	number = {2},
	journal = {ACM Transactions on Autonomous and Adaptive Systems},
	author = {Lewis, Peter R. and Esterle, Lukas and Chandra, Arjun and Rinner, Bernhard and Torresen, Jim and Yao, Xin},
	month = jun,
	year = {2015},
	pages = {1--30},
}

@inproceedings{pandey_hybrid_2020, 
	title = {Hybrid {Planning} {Using} {Learning} and {Model} {Checking} for {Autonomous} {Systems}},
	copyright = {https://ieeexplore.ieee.org/Xplorehelp/downloads/license-information/IEEE.html},
	language = {en},
	booktitle = {2020 {IEEE} {International} {Conference} on {Autonomic} {Computing} and {Self}-{Organizing} {Systems} ({ACSOS})},
	publisher = {IEEE},
	author = {Pandey, Ashutosh and Ruchkin, Ivan and Schmerl, Bradley and Garlan, David},
	month = aug,
	year = {2020},
	pages = {55--64},
}

@TechReport{UCAM-CL-TR-362,
  author =	 {Humphrys, Mark},
  title = 	 {{W-learning: competition among selfish Q-learners}},
  year = 	 1995,
  month = 	 apr, 
  institution =  {University of Cambridge, Computer Laboratory},
  doi = 	 {10.48456/tr-362},
  number = 	 {UCAM-CL-TR-362}
}

@article{chen_multiobjective_2021,
	title = {Multiobjective {Scheduling} {Strategy} {With} {Genetic} {Algorithm} and {Time}-{Enhanced} {A}* {Planning} for {Autonomous} {Parking} {Robotics} in {High}-{Density} {Unmanned} {Parking} {Lots}},
	volume = {26},
	copyright = {https://ieeexplore.ieee.org/Xplorehelp/downloads/license-information/IEEE.html},
	issn = {1083-4435, 1941-014X},
	language = {en},
	number = {3},
	journal = {IEEE/ASME Transactions on Mechatronics},
	author = {Chen, Guang and Hou, Jing and Dong, Jinhu and Li, Zhijun and Gu, Shangding and Zhang, Bo and Yu, Junwei and Knoll, Alois},
	month = jun,
	year = {2021},
	pages = {1547--1557},
}

@article{cheng_multi-objective_2020,
	title = {Multi-{Objective} {Genetic} {Algorithm}-{Based} {Autonomous} {Path} {Planning} for {Hinged}-{Tetro} {Reconfigurable} {Tiling} {Robot}},
	volume = {8},
	copyright = {https://creativecommons.org/licenses/by/4.0/legalcode},
	issn = {2169-3536},
	language = {en},
	journal = {IEEE Access},
	author = {Cheng, Ku Ping and Mohan, Rajesh Elara and Khanh Nhan, Nguyen Huu and Le, Anh Vu},
	year = {2020},
	pages = {121267--121284},
}

@article{zhang_multi-objective_2022,
	title = {Multi-objective particle swarm optimization with multi-mode collaboration based on reinforcement learning for path planning of unmanned air vehicles},
	volume = {250},
	issn = {09507051},
	language = {en},
	journal = {Knowledge-Based Systems},
	author = {Zhang, Xiangyin and Xia, Shuang and Li, Xiuzhi and Zhang, Tian},
	month = aug,
	year = {2022},
	pages = {109075},
}

@inproceedings{ramezani_human-centric_2024, 
	title = {Human-{Centric} {Aware} {UAV} {Trajectory} {Planning} in {Search} and {Rescue} {Missions} {Employing} {Multi}-{Objective} {Reinforcement} {Learning} with {AHP} and {Similarity}-{Based} {Experience} {Replay}*},
	copyright = {https://doi.org/10.15223/policy-029},
	language = {en},
	booktitle = {2024 {International} {Conference} on {Unmanned} {Aircraft} {Systems} ({ICUAS})},
	publisher = {IEEE},
	author = {Ramezani, Mahya and Atashgah, M.A. and Sanchez-Lopez, Jose Luis and Voos, Holger},
	month = jun,
	year = {2024},
	pages = {177--184},
}

@article{article_AtariDRL,
author       = {Volodymyr Mnih and
                  Koray Kavukcuoglu and
                  David Silver and
                  Alex Graves and
                  Ioannis Antonoglou and
                  Daan Wierstra and
                  Martin A. Riedmiller},
  title        = {Playing Atari with Deep Reinforcement Learning},
  journal      = {CoRR},
  volume       = {abs/1312.5602},
  year         = {2013},
  eprinttype    = {arXiv},
  eprint       = {1312.5602},
  bibsource    = {dblp computer science bibliography, https://dblp.org}
}

@inproceedings{cardozo2017peace,
  title={Peace COrP: Learning to solve conflicts between contexts},
  author={Cardozo, Nicol{\'a}s and Dusparic, Ivana and Castro, Jorge H},
  booktitle={Proceedings of the 9th ACM International Workshop on Context-Oriented Programming},
  pages={1--6},
  year={2017}
}

@article{kuvsic2021spatial,
  title={Spatial-temporal traffic flow control on motorways using distributed multi-agent reinforcement learning},
  author={Ku{\v{s}}i{\'c}, Kre{\v{s}}imir and Ivanjko, Edouard and Vrbani{\'c}, Filip and Greguri{\'c}, Martin and Dusparic, Ivana},
  journal={Mathematics},
  volume={9},
  number={23},
  pages={3081},
  year={2021},
  publisher={MDPI}
}

@inproceedings{ferreira_multi-objective_2017, 
	title = {Multi-objective reinforcement learning-based deep neural networks for cognitive space communications},
	language = {en},
	urldate = {2024-07-02},
	booktitle = {2017 {Cognitive} {Communications} for {Aerospace} {Applications} {Workshop} ({CCAA})},
	publisher = {IEEE},
	author = {Ferreira, Paulo Victor R. and Paffenroth, Randy and Wyglinski, Alexander M. and Hackett, Timothy M. and Bilen, Sven G. and Reinhart, Richard C. and Mortensen, Dale J.},
	month = jun,
	year = {2017},
	pages = {1--8},
}

@article{he_toward_2023,
	title = {Toward personalized decision making for autonomous vehicles: {A} constrained multi-objective reinforcement learning technique},
	volume = {156},
	language = {en},
	urldate = {2024-07-02},
	journal = {Transportation Research Part C: Emerging Technologies},
	author = {He, Xiangkun and Lv, Chen},
	month = nov,
	year = {2023},
	pages = {104352},
}

@article{wang_enhancing_2024,
	title = {Enhancing vehicle ride comfort through deep reinforcement learning with expert-guided soft-hard constraints and system characteristic considerations},
	volume = {59},
	language = {en},
	urldate = {2024-07-02},
	journal = {Advanced Engineering Informatics},
	author = {Wang, Cheng and Cui, Xiaoxian and Zhao, Shijie and Zhou, Xinran and Song, Yaqi and Wang, Yang and Guo, Konghui},
	month = jan,
	year = {2024},
	pages = {102328},
}
%

\end{document}